%% file: main.tex
\documentclass[10pt,twocolumn,letterpaper]{article}

\usepackage[pagenumbers]{cvpr} 


\definecolor{cvprblue}{rgb}{0.21,0.49,0.74}
\usepackage[pagebackref,breaklinks,colorlinks,allcolors=cvprblue]{hyperref}

\usepackage{tocloft}

\usepackage{xcolor}

\usepackage{multirow}
\newcommand{\best}[1]{\textbf{#1}}

\definecolor{Gray}{gray}{0.95}
\usepackage{colortbl}
\newcolumntype{a}{>{\columncolor{Gray}}c}
\newcommand{\mc}[2]{\multicolumn{#1}{c}{#2}}

\newcommand{\MOCA}{\text{MotionCanvas}\xspace}

\usepackage{enumitem}

\usepackage{animate}

\newcommand*{\affaddr}[1]{#1} 
\newcommand*{\affmark}[1][*]{\textsuperscript{#1}}
\newcommand*{\email}[1]{\texttt{#1}}
\usepackage[symbol]{footmisc}

\renewcommand{\thefootnote}{\fnsymbol{footnote}}
\newcommand\blfootnote[1]{
    \begingroup
    \renewcommand\thefootnote{}\footnote{#1} 
    \addtocounter{footnote}{-1}
    \endgroup
}
\def\paperID{*****} 
\def\confName{CVPR}
\def\confYear{2025}

\title{MotionCanvas:\\Cinematic Shot Design with Controllable Image-to-Video Generation}

\author{%
Jinbo Xing\affmark[1]$^{,*}$~~~~~~~~
Long Mai\affmark[2]~~~~~~~~
Cusuh Ham\affmark[2]~~~~~~~~
Jiahui Huang\affmark[2]~~~~~~~~
Aniruddha Mahapatra\affmark[2]\\
Chi-Wing Fu\affmark[1]~~~~~~~~~~~
Tien-Tsin Wong\affmark[3]~~~~~~~~~~~
Feng Liu\affmark[2]\\
\affaddr{\affmark[1]The Chinese University of Hong Kong~~~~~~~~~~}
\affaddr{\affmark[2]Adobe Research~~~~~~~~~~}
\affaddr{\affmark[3]Monash University}\\
\small\email{Project page: \url{https://motion-canvas25.github.io/}}
}

\begin{document}
\twocolumn[{
\renewcommand\twocolumn[1][]{#1}
\maketitle
\begin{center}
    \centering
    \captionsetup{type=figure}
    \animategraphics[autoplay,loop,width=\textwidth, trim=0 7 0 7, clip]{24}{Frames/teaser/}{0}{30}
    \vspace{-2em}
    \captionof{figure}{\MOCA offers comprehensive motion controls to animate a static image (the ``Inputs'' column) with various types of camera movements and object motions. Note the different camera movements across columns and object motions across rows. Please use \textbf{Adobe Acrobat Reader} for video playback.
    }
    \label{fig:teaser}
\end{center}
}]

\blfootnote{* Work done during an internship at Adobe Research.}

\begin{abstract}
This paper presents a method that allows users to design cinematic video shots in the context of image-to-video generation. Shot design, a critical aspect of filmmaking, involves meticulously planning both camera movements and object motions in a scene.
However, enabling intuitive shot design in modern image-to-video generation systems presents two main challenges: first, effectively capturing user intentions on the motion design, where both camera movements and scene-space object motions must be specified jointly; and second, representing motion information that can be effectively utilized by a video diffusion model to synthesize the image animations.
To address these challenges, we introduce \MOCA, a method that integrates user-driven controls into image-to-video (I2V) generation models, allowing users to control both object and camera motions in a scene-aware manner. 
By connecting insights from classical computer graphics and contemporary video generation techniques, we demonstrate the ability to achieve 3D-aware motion control in I2V synthesis without requiring costly 3D-related training data.
\MOCA enables users to intuitively depict scene-space motion intentions, and translates them into spatiotemporal motion-conditioning signals for video diffusion models.
We demonstrate the effectiveness of our method on a wide range of real-world image content and shot-design scenarios, highlighting its potential to enhance the creative workflows in digital content creation and adapt to various image and video editing applications.
\end{abstract}

\input{Sections/sec-1-intro}

\input{Sections/sec-2-related-work}

\input{Sections/sec-3-method}

\input{Sections/sec-4-applications}

\input{Sections/sec-5-evaluation}

\input{Sections/sec-6-conclusion}

{
    \small
    \bibliographystyle{ieeenat_fullname}
    \bibliography{main}
}
\clearpage
\input{Sections/X_supp}

\end{document}

%% file: Sections/sec-1-intro.tex
\section{Introduction}
\label{sec-1-intro}

Image animation, also known as image-to-video generation (I2V), empowers filmmakers and content creators with the ability to bring static images to life with dynamic motion. Image animation has attracted significant research efforts throughout the year, with various approaches explored to achieve realistic image animations~\cite{horrytour, xuanimating,liuinfinite, holynski2021animating}.
Recently, advances in video generation models have significantly transformed I2V synthesis~\cite{xing2023dynamicrafter,yang2024cogvideox}. These models utilize generative techniques to produce videos from text, opening up new creative possibilities. However, the limitations of pure I2V approaches become clear when considering the user-intent driven nature of dynamic content creation, as textual information alone falls short of capturing the intricate details of motion, leading to uncertainty and a lack of control in generating the animations.

In content creation and filmmaking, motion design of the shot is crucial. It involves planning the camera movements and object motions to create a cohesive visual narrative. 
For example, from a single input image in Fig.~\ref{fig:teaser} (leftmost column), different users may want to explore different camera movements (\eg, keeping the camera static, dollying backward, or orbiting while elevating) and object arrangements (\eg, moving the boys or the kites in particular directions)
to express their creative vision for the resulting footage.
However, current video generation models rely primarily on textual inputs for motion control, largely limiting the precision and expressiveness of the resulting animations. While some recent works~\cite{wang2024motionctrl,wu2025draganything} have attempted to incorporate camera and object motion controls into video diffusion models, they cannot fully address the challenges that come from the intertwining nature of camera and object motions, and hence struggle to interpret the ambiguous user intention. For instance, when the user drags the arm of a character in the image, it could mean panning the camera, moving the character, or just moving the arm.

In our work, we aim at enabling precise user control on the camera motion, object global motion (e.g., moving the character), and object local motion (e.g., moving the arm). A standard approach is to prepare video training data that have all required labels (e.g. camera extrinsic parameters, object motion, etc.). 
However, this requires labor-intensive and possibly unreliable labeling, and eventually limits the variety and quantity of video data for training~\cite{wang2024motionctrl,he2024cameractrl}. Instead, we propose to represent motions in ways that do not rely on such costly labels, but on simple 2D bounding boxes and 2D point trajectories that can be automatically and reliably estimated. This allows us to take almost any kind of video for training and enriches our generation variety. 

However, we still face one major challenge. 
Users tend to think and plan the motion in 3D scene space, but the video generation model is trained with 2D screen-space motion conditions. 
Moreover, users have to specify the motion plan on the 2D image canvas. In other words, we need an effective translation from the user motion design in scene space to control signals (bounding boxes and point trajectories) in 2D screen space. 
Inspired by the classical scene modeling in computer graphics, we propose the 
Motion Signal Translation module that allows users to plan the motion in 3D scene space, and then accurately interpret the user intention. 
This module can decompose the user motion into hierarchical transformation and rectify the potentially erroneous user input before feeding the interpreted 2D control signals to the video generation model.

We named 
our streamlined video synthesis system \textit{\MOCA},
allowing for joint manipulation of both camera and object motions in a scene-aware manner during the shot design and video synthesis processes.
As demonstrated in Fig.~\ref{fig:teaser}, our method allows users to generate video shots with different object motions while fixing a desired camera path (noting consistent camera motions across results in each column) and vice versa.
\MOCA leverages our dedicated Motion Signal Translation module during inference to connect high-level users' motion designs to the screen-space control signals that drive the video diffusion models for video synthesis.
To this end, we develop dedicated motion-conditioning mechanisms that guide DiT-based~\cite{peebles2023scalable} video diffusion models to follow control signals capturing camera and object motions.

\begin{itemize} 
\item We introduce cinematic shot design to the process of image-to-video synthesis. 
\item We present \MOCA, a streamlined video synthesis system for cinematic shot design, offering holistic motion controls for joint manipulation of camera and object motions in a scene-aware manner. 
\item In addition, we design dedicated motion conditioning mechanisms to guide DiT-based video diffusion models with control signals that capture camera and object motions. We couple that with a Motion Signal Translation module to translate the depicted scene-space motion intent into the screen-space conditioning signal for video generation. 
\item Evaluations on diverse real-world photos confirm the effectiveness of \MOCA for cinematic shot design, highlighting its potential for various creative applications. 
\end{itemize}

%% file: Sections/sec-2-related-work.tex
\section{Related Work}
\label{sec-2-related-work}

\noindent \textbf{Image Animation.} \
Early work drew heavily on classical graphics techniques, often aiming to reconstruct 3D geometry from a single 2D image and then re-render it at novel views. Notably, \emph{`Journey into the Picture'} established a pipeline for extracting planar layers and synthesizing minimal 3D scenes, while 3D Ken Burns~\cite{niklaus3d} harnessed estimated depth to produce convincing camera parallax, which is further improved in SynSin~\cite{wiles2020synsin}.
However, these methods are fundamentally limited to generating camera effects on static photographs without object motions. 

Beyond camera motions, another line of work animates only selective parts of an image, generally referred to as \emph{`cinematographs'}~\cite{chuang2005animating}. They manually define different types of motion for different subject classes. More recent works~\cite{mahapatra2023synthesizing, holynski2021animating, fan2022SLR} have employed deep networks to predict motion to generate cinemagraphs. However, these methods can only generate subtle motion for specific classes of elements like clouds, smoke, etc.

\input{Figures/fig-overall-pipeline}

\vspace*{1mm}
\noindent \textbf{Image-to-Video Generation.}  \
The field of video foundation models~\cite{blattmann2023stable,xing2023dynamicrafter,chen2023videocrafter1,videoworldsimulators2024,bar2024lumiere,polyak2024movie,yang2024cogvideox,hacohen2024ltx} has seen significant advances in recent years. These methods demonstrate great abilities to generate highly realistic videos from a single image and a user-specified text prompt. However, they typically offer limited or no user control over the motion in the generated videos, 
thereby limiting their applicability. 

\vspace*{1mm}
\noindent \textbf{Controllable Video Generation.}  \
Existing works~\cite{wang2024boximator,wu2025draganything,xing2024make,wang2024motionctrl} typically offer one of these three forms of user-guided control: 
\textbf{(1) Bounding Box Control}~\cite{wang2024boximator,ma2024trailblazer,huang2023factor,li2023trackdiffusion,wu2024motionbooth}.
They allow users to control object motion by drawing a sequence of bounding boxes starting from the object position. Since these works do not provide controllability for camera motion, the bounding box control alone inevitably introduces ambiguity,~\eg, it is not clear to move objects or the whole scene to follow the bbox positions.
\textbf{(2) Point Trajectory Control}~\cite{wu2025draganything,niu2025mofa,wang2024motionctrl,mou2024revideo,qiu2024freetraj}.
While these methods provide decent drag-based motion control using point trajectories, they fail to account for object motion in a 3D-aware manner and are often restricted to only a limited number of trajectories.
\textbf{(3) Camera Control}~\cite{wang2024motionctrl,he2024cameractrl,yu2024viewcrafter,wang2024akira,bahmani2024ac3d}.
This line of work provides camera motion control in video generation with explicit 3D camera parameters.  However, the training dataset for these methods is restricted to mostly static scenes and curated domains, so they can only produce few object motions and lack generalizability.

In contrast, our approach provides joint object-level control (via bounding boxes or point trajectories) and camera-level control in a unified framework. 
Moreover, none of the prior methods mentioned integrate scene-aware object control, due to a lack of 3D training data. 
Instead, 
we train solely on 2D signals (\eg, point trajectories) and allow 3D-aware controls with depth-based synthesis.

%% file: Figures/fig-overall-pipeline.tex
\begin{figure*}[t]
    \centering
    \includegraphics[width=0.97\textwidth]{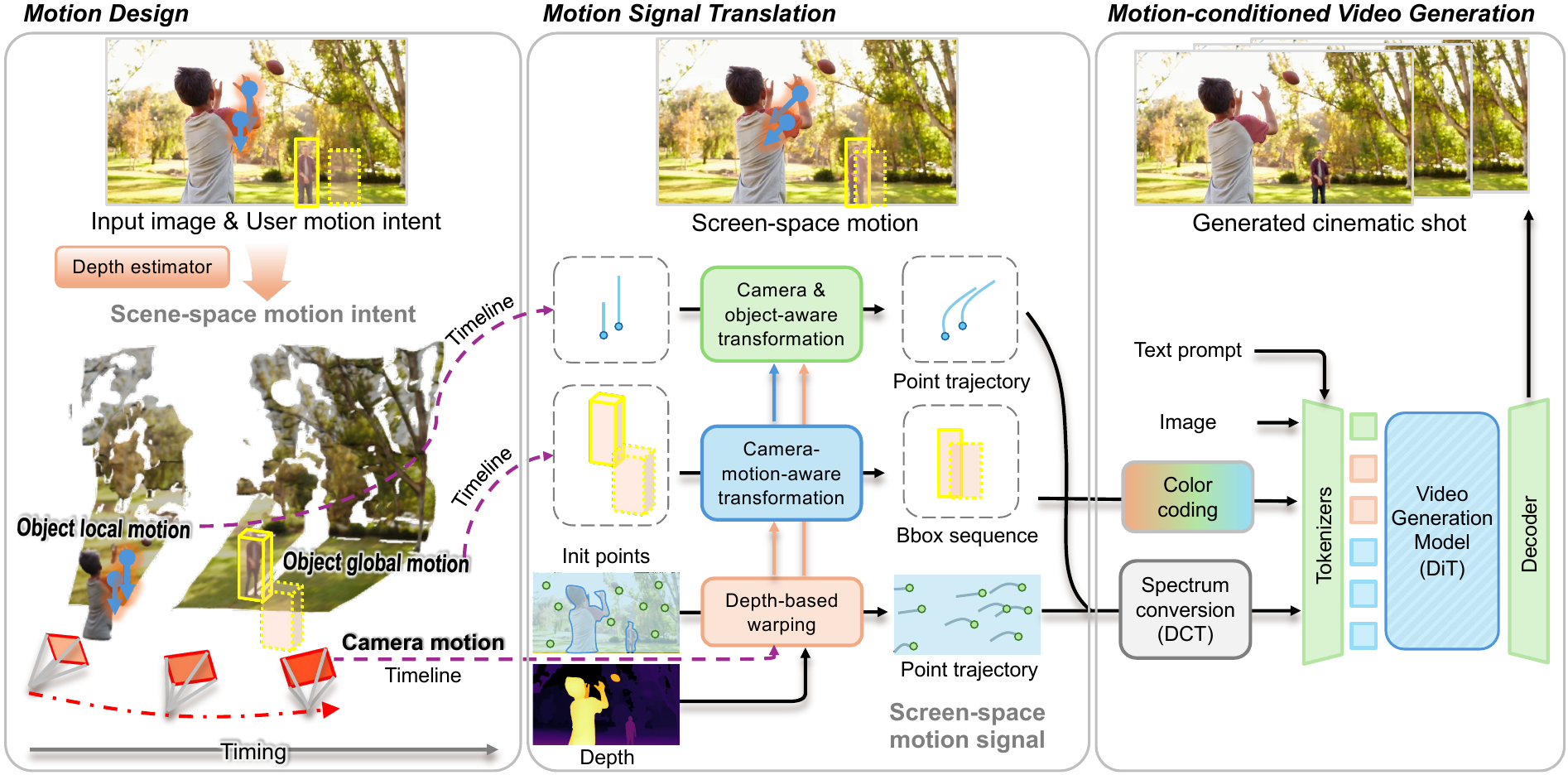} 
    \caption{Overview of \textit{MotionCanvas}. Given an input image and high-level scene-space motion intent, MotionCanvas decomposes and translates the motion (camera and object motion with their timing) into screen space by leveraging the depth-based synthesis and hierarchical transformation with the Motion Signal Translation module. These screen-space motion signals are subsequently passed to a video generation model to produce the final cinematic shots.
    }
    \label{fig:overall-pipeline}
\end{figure*}

%% file: Sections/sec-3-method.tex
\section{\MOCA}
\label{sec-3-method}

Our method animates still images into short videos, reflecting user's motion design intentions. As Fig.~\ref{fig:overall-pipeline} illustrates, 
\MOCA comprises three main components: (1) the motion design module to capture diverse scene-aware motion intentions, (2) the translation module to convert these intentions into screen-space motion signals, and (3) the motion-conditioned video generation model.

\subsection{Motion Design Module -- Capturing User Intents}
\label{sec-3.1-motion-design}
We leverage the input image as the canvas for motion design, establishing a starting scene on which the motion designs are grounded. This setting enables 3D-scene awareness in the motion design,
capturing spatial relationships between objects, the camera, and the scene. Our unified interface facilitates individual control over camera motion, object global and local motion, and their timing.

\textbf{Camera Motion Control with 3D Camera Paths.}
We define camera motion using standard pinhole camera parameters, specifying a 3D camera path as a sequence of extrinsic $E_l$ and intrinsic $K_l$ parameters for each frame $l$: $(E_l, K_l)_{l=1}^{L}$, where $E_l\in \mathbb{R}^{3\times4}$ and $K_l \in \mathbb{R}^{3\times 3}$.
To design the camera path, users can specify the camera poses (\ie, the translation vector and rotation angles with respect to the initial view) at key moments. The full camera path can be obtained via interpolation.
To make the camera path specification process more intuitive for users, our system also lets users specify and mix $M$ base motion patterns (\eg, panning, dolly) along with the corresponding direction and speed. Our method then converts the specified motion patterns into 3D camera paths (see section~\ref{sec:ui} in the supp.).

\textbf{Object Global Motion Control with Scene-anchored Bounding-boxes.}
Controlling where the objects move to within the scene is crucial in designing video shots. 
We argue that such global object control should be defined in a scene-aware manner, in which object positioning is grounded to positions in the underlying 3D scene.
To this end, we enable scene-anchored bounding box (bbox) placement by minimally specifying start and end boxes, as well as (optionally) intermediate key boxes, on the input image. 
By anchoring the bbox placements to the fixed view established by the input image, users can depict the imagined target locations via adjusting not only the position but also the scale and shape of the boxes. This scene-aware bbox placement provides intuitive control over object position, scale, pose, and relative distance to the camera.
From the provided key positions and intended duration of the output video, we generate smooth box trajectories via Catmull-Rom spline interpolation.

\textbf{Object Local Motion Control with Point Tracing.}
While global object motion defines how objects change their position within the scene and is our main focus in the shot design process, local object motion -- depicting in-location movement of objects (\eg, raising arms, a rotating head) -- can also enrich the shot design experience with added details and realism.
Inspired by the recent success of drag-based editing~\cite{mou2024revideo}, we use sparse point trajectory to depict local motion. As local motion often involves complex geometric relationships and deformations, sparse point trajectories offer a flexible way to define and manipulate such motion.

\textbf{Timing Control}.
Timing control of object and camera motions enables coordinated design, enhancing narrative flow and visual coherence. Our system naturally supports this by allowing users to assign timelines directly along the motion trajectories.

\subsection{Motion Signal Translation Module}

While motion intent is best designed in the 3D-aware scene-centric manner, video generation models are often most effectively trained for motion conditioning with 2D screen-space, frame-centric data that mix all motion types together after view-dependent projections. 
This discrepancy arises due to the challenging nature of extracting reliable 3D information such as camera motions and 3D object trackings in general videos at large scales~\cite{zhou2018stereo}. 
To address this challenge, instead of aiming to devise a video generation model that directly handles the scene-space motion information, our key idea is to translate the scene-space motion design obtained from Section~\ref{sec-3.1-motion-design} into spatiotemporally grounded screen-space motion signals that can be reliably extracted from in-the-wild videos. 

\textbf{Camera Movement via Point Tracking.}
We seek a form of screen-space motion signal that (1) can be robustly extracted from general videos, and (2) encodes detailed information about the camera movement throughout the videos.
Research on human visual perception~\cite{epstein1995perception, thompson2011perception} offers the important insight that egocentric motion can be reliably recovered by a sparse set of scene point trackings projected onto the image plane. This insight has been widely applied in computer vision for camera pose estimation~\cite{hartley2003vision} and SLAM~\cite{taketomi2017visual}.
Inspired by this insight, we represent camera movement using point tracking. Note this information can be robustly extracted from real videos~\cite{cho2025local}. 

At inference time, we convert 3D camera paths to 2D point tracks by randomly sampling a set of points on the input image. To focus on points belonging to static background, which better reflect the camera movement, we exclude regions from likely-moving objects estimated from a YOLOv11~\cite{khanam2024yolov11}-generated mask. 
We then employ an off-the-shelf monocular depth estimator~\cite{wang2024moge} to obtain intrinsic camera parameters and a depth map. Lastly, we warp these points based on the 3D camera path and depth to create corresponding 2D screen-space tracks.

\textbf{Scene-aware Object Motion via Bounding-box Trajectories.}
User-defined scene-space bbox trajectories reflect motion intent but are distorted by camera movement and perspective when projected onto the screen. 
We aim to convert such scene-anchored object bboxes to screen space in a way that resembles the object bbox sequence typically extracted from real videos. 
We first lift the scene-space bbox to 2.5D, using camera pose and depth to continuously reproject it into screen space across frames. The initial bbox's depth is assigned the average depth within its SAM2~\cite{ravi2024sam}-generated semantic mask. Subsequent bboxes use either (1) a reference depth from a point in the scene (e.g., the ground plane) or (2) a depth from perspective consistency (a growing bbox implies movement toward camera).
Using the assigned depth and camera pose transformations, the 2.5D bboxes $b_\text{scene}^l$ at time $l$ are projected into screen space $b_\text{screen}^l$ with calibrated locations and sizes:
\begin{equation}
    b_\text{screen}^l = \mathcal{T}_\text{camera}^l(b_\text{scene}^l),
\end{equation}
where $\mathcal{T}_\text{camera}^l(\cdot)$ denotes the camera motion transformation.

\textbf{Object Local Motion via Point Trajectory Decomposition.}
As we aim to utilize the scene-anchored point trajectory to depict object local motion, our focus lies on converting each scene-anchored control point $p_\text{scene}^l$ (obtained by lifting the control point to scene space using the corresponding object-bbox's depth computed previously) into the corresponding screen space $p_\text{screen}^l$. This involves transformations considering both camera and global motion:
\begin{equation}
    p_\text{screen}^l = \mathcal{T}_\text{camera}^l(\mathcal{T}_\text{global}^l(p_\text{scene}^l)),
\end{equation}
where $\mathcal{T}_\text{global}^l(\cdot)$ represents object global motion transformation. Assuming negligible depth changes during local motion relative to global motion---a reasonable assumption as local motion often occurs within a similar depth plane (\eg, a waving hand)---we assign all local motion points the same depth as their initial positions. This simplifies the transformation based on camera motion.

\subsection{Motion-conditioned Video Generation}
Video diffusion models have emerged as the dominant paradigm for video generation.
We build our motion-conditioned video generation model upon a pre-trained Diffusion Transformer (DiT)-based~\cite{peebles2023scalable} I2V model.
This model is an internally-developed standard adaptation of DiT to video generation, similar to existing open-source adaptions~\cite{opensora,yang2024cogvideox}.
We adapt the model to our motion-conditioned generation problem by fine-tuning it with the screen-space motion conditions.

\input{Figures/fig-method-motion-conditioning}

\textbf{Point-Trajectory Conditioning.}
We represent $N$ point trajectories by encoding each into a compact set of Discrete Cosine Transform (DCT) coefficients. Due to the low temporal frequency nature of point tracking, we can compress long trajectories into $K$ coefficients ($K$=10 in our experiments), with the DC component encoding the initial position, grounding the trajectory's start point explicitly. This compact representation, $C_\text{traj}\in\mathbb{R}^{N\times K \times 2}$, offers two advantages: (1) it simplifies motion data handling, allows flexible handling of varying trajectory densities, and improves efficiency through pre-computation; and (2) it integrates seamlessly into the DiT framework via in-context conditioning. Each trajectory is treated as an individual token, with its embedding derived from the DCT coefficients.

\textbf{Bbox-Sequence Conditioning.}
Bounding boxes convey complex spatial information (position, shape, pose) and are injected separately from trajectories to distinguish between camera and object global motion. 
We encode bbox sequences as spatiotemporal embeddings by first rasterizing them into unique color-coded masks, forming an RGB sequence $C_\text{bbox}^\text{RGB}\in \mathbb{R}^{L\times H\times W\times 3}$.
This sequence is then encoded into spatiotemporal embeddings $C_\text{bbox}\in \mathbb{R}^{L'\times H'\times W'\times C}$ using the same pre-trained video autoencoder (3D-VAE) from the base DiT model. These embeddings are patchified into tokens and added to the noisy video latent tokens, as illustrated in Fig.~\ref{fig:motion-conditioning}.

\textbf{Model Training.}
Since we adopt a latent diffusion model, all RGB-space data are compressed into tokens within the latent space via a 3D-VAE encoder. The motion-conditioned model is optimized using the flow-matching loss~\cite{lipmanflow,liu2023instaflow}. Specifically, denoting the ground-truth video latents as $X^1$ and noise as $X^0\sim\mathcal{N}(0,1)$, the noisy input is produced by linear interpolation $X^t=t X^1+(1-t)X^0$ at timestep $t$. The model $v_{\theta}$ predicts the velocity $V^t=\frac{d}{dt}X^t=X^1-X^0$. The training objective is
\begin{equation}
    \min _{\theta} \mathbb{E}_{t,X^0,X^1} \left[\Vert V^t -v_{\theta}(X^t,t \,|\, C_\text{img},C_\text{traj},C_\text{bbox},C_\text{txt})   \Vert_2^2 \right],
\end{equation}
where $C_\text{img}$, $C_\text{traj}$, $C_\text{bbox}$, $C_\text{txt}$ denote the input image, point trajectory, bbox, and text prompt, respectively.
Note we also retain the text conditioning to offer additional flexibility in controlling video content changes. After training, the clean video latent tokens $\hat{X^1}$ can be generated conditioning on the input image, text prompt, and screen-space motion signals, and then decoded back to RGB frames.

\input{Figures/fig-app-shot-design}

\subsection{Generating Variable-length Videos via Auto-regression}

Generating variable-length videos is beneficial for cinematic storytelling. We achieve this through auto-regressive generation, which is more computationally efficient than modeling long videos directly and reflects the fact that complex video shots are often composed of short, simple shots stitched in a sequence.
While our image-to-video framework naturally supports training-free long video generation in an auto-regressive manner, we found that this often results in noticeable motion discontinuities, as a single conditional image lacks sufficient temporal motion information.
To address this, we train MotionCanvas$_\text{AR}$ with an additional conditioning on short video clips, $C_\text{vid}$ (16 frames). This overlapping short-clip strategy conditions each generation step on the previous spatiotemporal context, resulting in natural transitions. During inference, the model generates videos of arbitrary length with independent motion control per generation iteration. 
To further refine the input motion signals and align them with the training setup, we recompute the screen-space motion signals by combining the user’s intent with back-traced motions.
This approach ensures smoother and more consistent motion generation (see section~\ref{sec:ar} in the supplement).
\input{Figures/fig-app-shot-design-longvid}

%% file: Figures/fig-method-motion-conditioning.tex
\begin{figure}[t]
    \centering
    \includegraphics[width=1\linewidth]{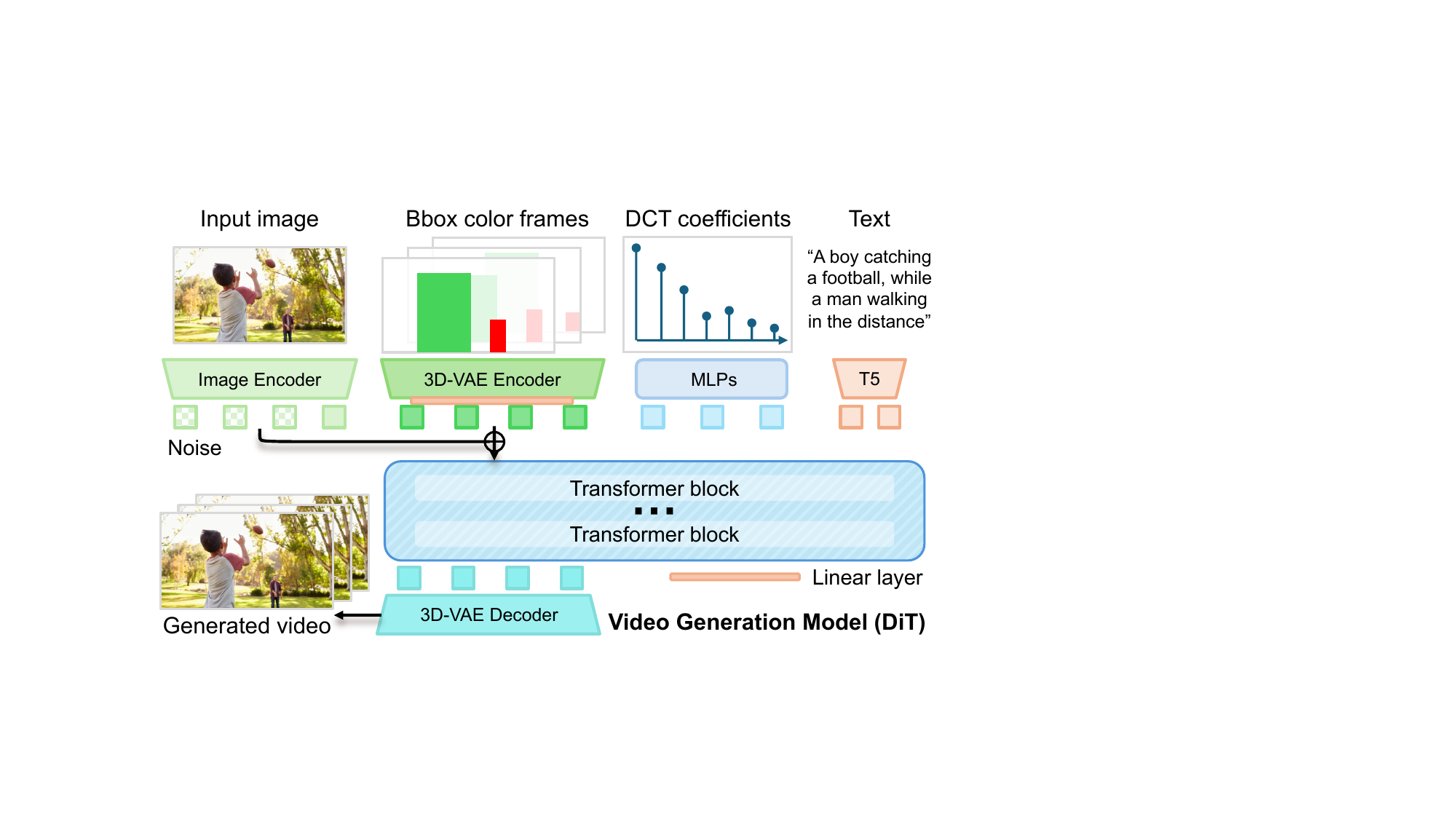}
    \caption{Illustration of our motion-conditioned video generation model. The input image and bbox color frames are tokenized via a 3D-VAE encoder and then summed. The resultant tokens are concatenated with other conditional tokens, and fed into the DiT-based video generation model.}  
    \label{fig:motion-conditioning}
\end{figure}

%% file: Figures/fig-app-shot-design.tex
\begin{figure*}[htbp]
    \centering
    \animategraphics[autoplay,loop,width=1.0\linewidth, trim=0 0 0 0, clip]{24}{Frames/app-shot-design/}{0}{30}
    \caption{Shot design generated by our \MOCA under various types of joint camera and object motion controls.}
    \label{fig:app-shot-design}
\end{figure*}

%% file: Figures/fig-app-shot-design-longvid.tex
\begin{figure*}[t]
    \centering
    \animategraphics[autoplay,loop,width=1.0\linewidth, trim=0 0 0 0, clip]{24}{Frames/app-longvid/}{0}{158} 
    \caption{Long videos with the same complex sequences of camera motion while different object motion controls in each case generated by our \MOCA.}
    \label{fig:app-shot-design-longvid}
\end{figure*}

%% file: Sections/sec-4-applications.tex
\section{Applications}
\label{sec-4-applications}

\MOCA allows for flexible controls of both camera and object motion in a scene-aware manner. This enables our main application of cinematic shot design framework, allowing users to interactively manage key motion aspects of a shot. 
Additionally, the flexibility of our motion representations makes it possible to naturally apply our framework on a variety of simple video-based editing tasks.

\subsection{Shot Design with Joint Camera and Object Control}

As illustrated in Fig.~\ref{fig:app-shot-design}, our framework enables precise and independent control of object and camera motion in a scene-aware manner, allowing for the design of highly dynamic and visually compelling shots while closely following the provided motion design. 

In Fig.~\ref{fig:app-shot-design}, it is worth noting that in both examples, the results in each column follow the same camera motion while the object motions change according to the corresponding specified object controls. By placing the bounding box in a scene-aware manner, users can achieve various scene-space effects. For example, this makes it possible to induce the car in the bottom example to stay still (first row) or move forward (second row) and backward (third row) on the road. Importantly, such scene-anchored motion is preserved while the camera motion changes independently. 
This highlights the importance of scene-aware object motion controls.

\textbf{Long Videos with Complex Trajectories.}
To generate extended videos featuring complex camera and object motions, our framework employs a ``specification-generation'' loop. This approach allows users to define motion signals for each segment and subsequently generate video chunks in an auto-regressive manner. Inspired by animation workflows~\cite{xing2024tooncrafter,tang2025generativeaicelanimationsurvey}, \MOCA combines keyframing and interpolation to create intricate motion paths. Specifically, users can set keyframes for object and camera motions, then the system interpolates between these keyframes to produce smooth and coherent trajectories.

As demonstrated in Fig.~\ref{fig:app-shot-design-longvid}, our method can generate long videos with complex sequences of camera motion controls. We show two video results for each input image, resulting from the same camera controls (note that almost identical camera motions are generated for those two) while intentionally controlling different object motions.

\input{Figures/fig-app-shot-design-local-motion}

\subsection{Object Local Motion Control}

\MOCA also enables controlling object local motion to potentially support drag-based editing and generation. 
Users can define local object motions by directly specifying drag-based trajectories within the object's own coordinates. These point trajectories are then transformed to the appropriate screen-space point trajectories for conditioning the video generation model, taking into account both the camera and the object global motion. 
As illustrated in Fig.~\ref{fig:app-shot-design-local-motion}, our method can generate diverse and fine-grained local motions, making it possible to generate different variations of object movements (\eg, the different ways the arms of the baby move).

In addition, thanks to our dedicated motion translation module that accounts for the coordination between local motion and camera motion, as well as object global motion,
consistent object local motion control can be achieved with different camera and object dynamics (Fig.~\ref{fig:app-shot-design-local-motion} (bottom)). 
This opens the possibility of incorporating local object motion control into the shot design framework described above.

\input{Figures/fig-additional-apps}

\subsection{Additional Applications: Simple Video-based Editing}

\textbf{Motion Transfer.}
Our method can be adapted to perform motion transfer from a source video to an input image that shares a structural similarity with the initial frame. By leveraging the versatile screen-space conditioning representation, our framework effectively captures and transfers both object and camera motions, even for cases that involve 3D transform, without the need for explicit 3D camera pose extraction. As shown in Fig.~\ref{fig:additional-apps}, the rotating movement of the apple can be transferred to rotate the lion's head.

\textbf{Video Editing.}
The concept of motion transfer can be extended to facilitate video editing, where the input image is derived from the first frame through image editing~\cite{brooks2023instructpix2pix}. Utilizing the versatile screen-space conditioning representation, our method propagates extracted object and camera motions to the derived image, ensuring consistent and realistic dynamics, similar to~\cite{liu2024generative}. Fig.~\ref{fig:additional-apps} shows two examples where the edits performed on the initial frame are propagated using the motion signals extracted from the original video, resulting in a fully edited video.
%

\input{Figures/fig-eval-camera-control}

\input{Figures/fig-eval-object-control}


%% file: Figures/fig-app-shot-design-local-motion.tex

\begin{figure*}[t]
    \centering
    \animategraphics[autoplay,loop,width=0.95\linewidth, trim=0 0 0 0, clip]{24}{Frames/app-local-motion/}{0}{30} 
    \caption{Generated videos with diverse and fine-grained local motion controls (upper), and in coordination with camera motion control (bottom).}
    \label{fig:app-shot-design-local-motion}
\end{figure*}

%% file: Figures/fig-additional-apps.tex
\begin{figure*}[t]
    \centering
    \includegraphics[width=1\linewidth]{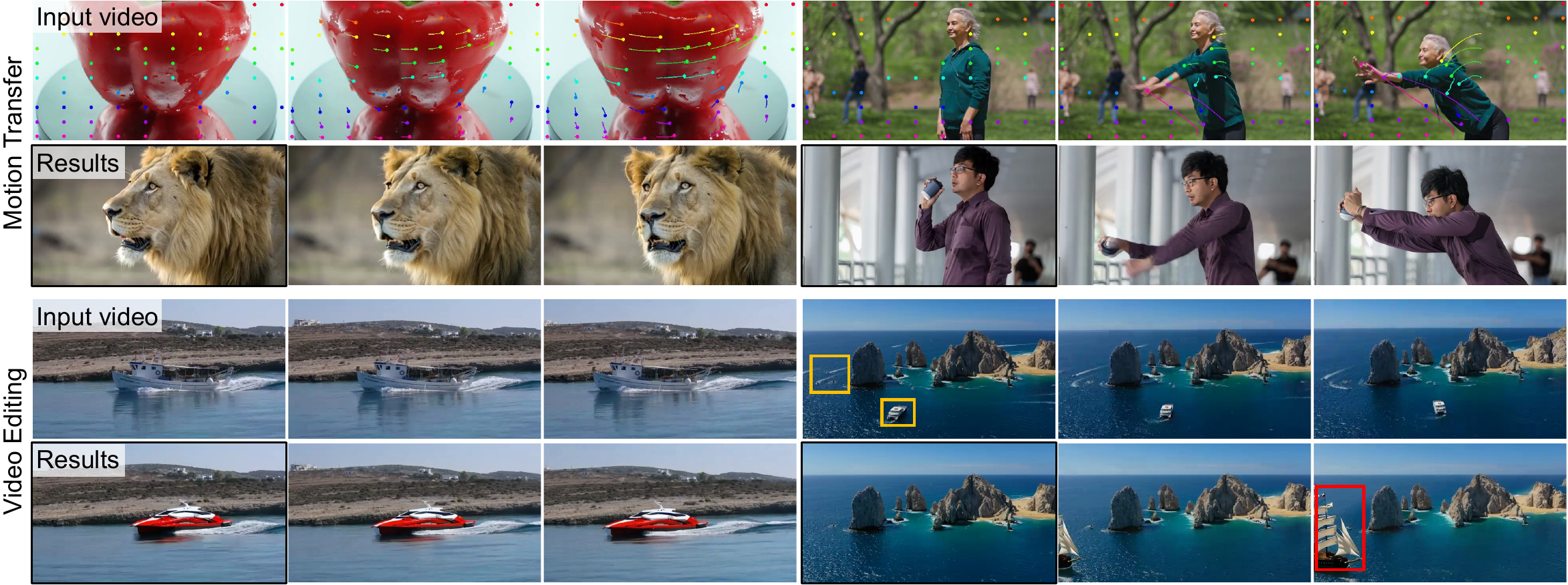} 
    \caption{Results when our method is applied for: (upper) motion transfer, and (bottom) video editing for changing objects, adding and removing objects.}
    \label{fig:additional-apps}
\end{figure*}

%% file: Figures/fig-eval-camera-control.tex
\begin{figure*}[t]
    \centering
    \animategraphics[autoplay,loop,width=1.0\linewidth, trim=0 0 25 0, clip]{24}{Frames/eval-camera-compare/}{0}{30}
    \caption{Camera motion control comparison. Compared to existing baselines, our method performs better at following the intended camera motion, especially for complex shot type such as the ``Dolly-Zoom" effect (second example).}
    \label{fig:eval-camera-control}
\end{figure*}

%% file: Figures/fig-eval-object-control.tex
\begin{figure*}[htbp]
\centering
\includegraphics[width=1.0\linewidth]{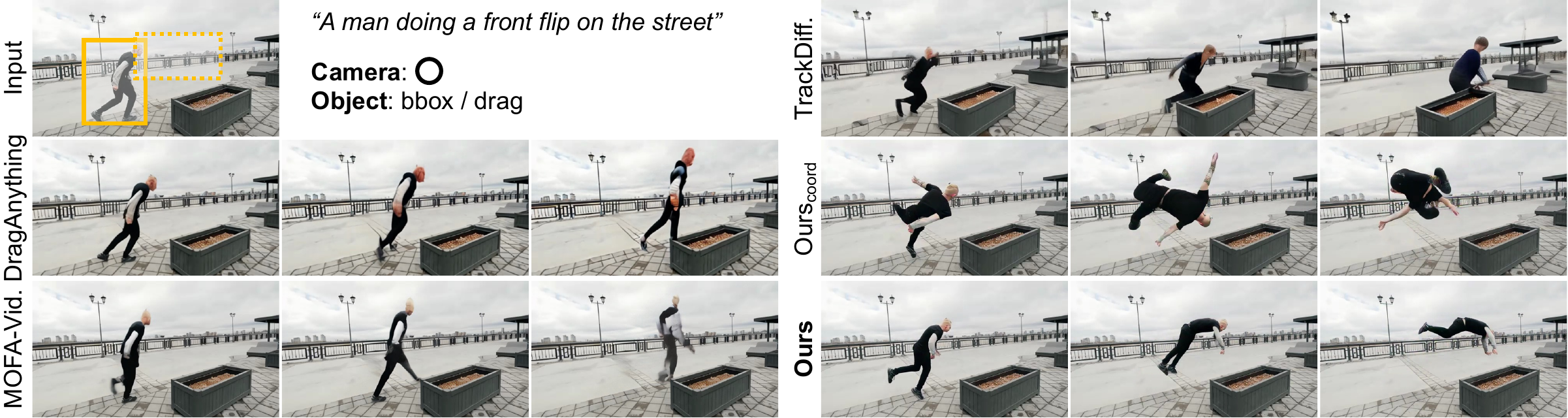} 
    \caption{Visual comparison of the resulatant videos from DragAnything, MOFA-Video, TrackDiffusion, Ours$_\text{coord}$, and our MotionCanvas.}
    \label{fig:eval-object-control}
\end{figure*}

%% file: Sections/sec-5-evaluation.tex
\section{Experiments}
\label{sec-5-evaluation}

\subsection{Implementation Details}

\input{Tables/tab-camera-motion}

\noindent\textbf{Data.} We collected $\sim$1.1M high-quality videos from our internal dataset. We extracted bounding boxes from the videos by performing panoptic segmentation with DEVA~\cite{cheng2023tracking}, followed by fitting the bounding boxes to the extracted masks. We compute sparse point tracking annotations by chaining optical flow from RAFT~\cite{teed2020raft}. To ensure reliable motion data, we established a threshold for valid tracking length. 
We also filtered out a subset of videos based on keywords such as $\texttt{vector}$, $\texttt{animation}$ to focus on natural video data. Bounding boxes were further refined using thresholds for adjacent-frame Intersection over Union (IoU), size change ratio, position change (Euclidean distance), and the relevance of the associated object to our list of moving objects. In the end, we obtained around 600K videos with good motion quality and high-fidelity annotations. During training, we randomly selected $N$ point trajectories with an 80\% probability, where $N \sim \mathcal{U}(0, 100)$. Additionally, there was a 10\% probability of selecting points exclusively from moving object regions, and another 10\% probability for points from non-moving object regions.

\noindent\textbf{Model.} Our video generation module is fine-tuned from a pre-trained image-to-video DiT model for 100K steps, utilizing a batch size of 256 and the AdamW optimizer~\cite{loshchilov2017decoupled} with a learning rate of $1 \times 10^{-5}$ and a weight decay of 0.1. The training primarily involved videos with 32 and 64 frames, sampled at 12 and 24 FPS with a resolution of 640$\times$352. During inference, we apply classifier-free guidance for text condition~\cite{ho2022classifier}.

\input{Tables/tab-object-global-motion}

\subsection{Camera Motion Control Quality}

We adopt rotation error (RotErr.), translation error (TransErr.), and CamMC as metrics, following~\cite{he2024cameractrl,wang2024motionctrl}. Additionally, we compute the Fréchet Inception Distance (FID)~\cite{heusel2017gans} and the Fréchet Video Distance (FVD)~\cite{unterthiner2019fvd} to assess the quality of the generated videos. These metrics are computed on 1K videos randomly sampled from the RealEstate-10K~\cite{zhou2018stereo} test set (@640$\times$352 with 14 frames).

We compare our method with two state-of-the-art camera-motion-controlled image-to-video methods: MotionCtrl~\cite{wang2024motionctrl} and CameraCtrl~\cite{he2024cameractrl}. The quantitative results are presented in Table~\ref{tab:quan_compare_camera}. Note that both MotionCtrl and CameraCtrl were trained on the RealEstate10K training set, which contains videos within the same domain as the test set. Nevertheless, our method outperforms them across all metrics in a zero-shot setting. 

The visual comparison in Fig.~\ref{fig:eval-camera-control} illustrates that the motion generated by MotionCtrl and CameraCtrl exhibits lower quality, primarily due to their reliance on training with video datasets (RealEstate10K) that include 3D camera pose labels, which lack diversity and consist solely of static scenes. Furthermore, our method allows for the control of intrinsic parameters, enabling the production of more advanced cinematic shots, such as dolly zoom (see Fig.~\ref{fig:eval-camera-control} (right)) which is difficult to achieve with existing methods.

\input{Tables/tab-user-study}

\subsection{3D-Aware Object Motion Control Quality}

Following~\cite{wu2025draganything}, we compute ObjMC and FID on the VIPSeg~\cite{miao2022large} filtered validation set, which includes 116 samples after excluding videos without moving objects (@640$\times$352 with 14 frames). We compare with DragAnything~\cite{wu2025draganything}, MOFA-Video~\cite{niu2025mofa}, and TrackDiffusion~\cite{li2023trackdiffusion}, with the quantitative results reported in Table~\ref{tab:quan_compare_object}. 
Our method outperforms the other baselines in both control accuracy (ObjMC) and frame quality (FID), as further evidenced in Fig.~\ref{fig:eval-object-control}. The explicit warping in DragAnything and MOFA-Video introduces object distortion while the reliance on Euclidean coordinates in TrackDiffusion hinders convergency, resulting in inaccuracy. By incorporating a spatiotemporal representation for bbox, our method enables the precise object motion controls (\eg, position, size, and pose).

\input{Figures/fig-eval-shot-design}

\subsection{Joint Camera and Object Control}

We conducted a user study (detailed in section~\ref{sec:user} in the supplement) to evaluate the perceptual quality of joint camera and object motion control in a 3D-scene-aware context. We compared our method with drag-based I2V methods: DragAnything~\cite{wu2025draganything} and MOFA-Video~\cite{niu2025mofa}. 
Note that none of the existing methods are designed for 3D-aware control, thus we directly take scene-space point trajectories as input to the baselines, following their original setting.
In addition to the point trajectory for object local motion control, we provided point trajectories from bounding box sequences and depth-based warping for global motion control of both object and camera. 
Participants were asked to select the best result based on motion adherence, motion quality, and frame fidelity. The statistics from the responses of 35 participants are summarized in Table~\ref{tab:user_study}. Our method consistently outperformed the competitors across all evaluated aspects. The visual results are presented in Fig.~\ref{fig:eval-shot-design}, where both baselines fail to jointly capture complex object global motion (\ie, the body's movement), local motion (\ie, putting down hands), and camera motion in a 3D-aware manner. In contrast, our \MOCA generates motions following all types of controls, thanks to its unified framework and design of motion representations.


\subsection{Ablation Studies}

\input{Tables/tab-ablation-study}

\input{Figures/fig-eval-camera-ablation}

\textbf{Camera Motion Representation.} We constructed several baselines to investigate the effectiveness of our camera motion representation: a Gaussian map (a 2D Gaussian blurred sparse optical-flow map), Plucker embedding~\cite{he2024cameractrl,xu2024camco,bahmani2024vd3d}, and our proposed DCT-coefficient-based trajectory encoding. The quantitative comparison is presented in Table~\ref{tab:ablation_study_camera}. The Gaussian map variant is inferior in accurate camera control due to its inherent ambiguity (especially under more dense controls), tending to generate static camera motions (high FVD). It is noteworthy that the Plucker embedding variant requires training on a video dataset with 3D camera pose labels, (\ie, RealEstate10K training set, following~\cite{he2024cameractrl}). It performs well on this in-domain static test set, but fails to generate object motions (Fig.~\ref{fig:eval-camera-ablation} `cat') and lacks generalizability. In addition, our trajectory coding is highly efficient, introducing only a few coefficient tokens while delivering robust performance for camera intrinsic and extrinsic controls.

\noindent\textbf{Bounding Box Conditioning.} We further evaluated our bounding-box conditioning. We applied an alternative conditioning design proposed in~\cite{wang2024boximator} that concatenates bounding-box coordinates to visual tokens (Ours$_\text{coord}$). The results in the last two columns of Table~\ref{tab:quan_compare_object} demonstrate the superiority of our spatiotemporal color-coding map conditioning. The difficulty of fusing Euclidean coordinate tokens with visual tokens leads to low ObjMC.

%% file: Tables/tab-camera-motion.tex
\begin{table}[t]
    \caption{Quantitative comparison with state-of-the-art methods on the RealEstate10K test set (1K).
  $^*$ denotes zero-shot performance.}
  \label{tab:quan_compare_camera}
\resizebox{\linewidth}{!}{
  \centering
  \begin{tabular}{lcccccc}
  \toprule
    Method  &RotErr$\downarrow$ &TransErr$\downarrow$ &CamMC$\downarrow$&FVD$\downarrow$ & FID$\downarrow$\\
    \midrule
MotionCtrl  & 0.8460& 0.2567& 1.2455& 48.03&11.34\\
 CameraCtrl  & 0.6355& 0.2332& 0.9532& 39.46& 13.14\\
 \rowcolor{Gray}
 Ours$^*$ & \best{0.6334} &\best{0.2188} & \best{0.9453}& \best{34.09}& \best{7.60}\\
  \bottomrule
  \end{tabular}
}
\end{table}

%% file: Tables/tab-object-global-motion.tex
\begin{table}[t]
  \caption{Quantitative comparison for object motion control on VIPSeg.}
  \label{tab:quan_compare_object}
\resizebox{\linewidth}{!}{
  \setlength{\tabcolsep}{2.1pt}
  \centering
  \begin{tabular}{lccc|ca}
  \toprule
    Metric & DragAnything &MOFA-Video &TrackDiffusion &Ours$_\text{coord}$ & \mc{1}{\multirow{1}{*}{Ours$_\text{map}$}} \\
    \midrule
ObjMC$\downarrow$  & 32.37& 35.94& 30.49&47.73 &\best{25.72}\\
 FID$\downarrow$ & 64.32& 54.58& 58.08& 46.27&\best{42.47}\\
  \bottomrule
  \end{tabular}
}
\end{table}

%% file: Tables/tab-user-study.tex
\begin{table}[t]
  \caption{User study statistics of the preference rate for motion adherence, motion quality, and frame fidelity.}
  \label{tab:user_study}
\resizebox{\linewidth}{!}{
  \setlength{\tabcolsep}{2.9pt}
  \centering
  \begin{tabular}{lccc}
  \toprule
    Method  & Motion Adherence$\uparrow$ & Motion Quality$\uparrow$& Frame Fidelity$\uparrow$\\
    \midrule
DragAnything  & 14.29\%& 10.10\%& 9.90\%\\
 MOFA-Video  & 10.48\%& 10.86\%& 12.95\%\\
 \rowcolor{Gray}
 Ours &75.24\%& 79.05\% & 77.14\%\\
  \bottomrule
  \end{tabular}
}
\end{table}

%% file: Figures/fig-eval-shot-design.tex
\begin{figure}[t]
    \centering
    \animategraphics[autoplay,loop,width=\linewidth, trim=0 3 0 5, clip]{24}{Frames/eval-user-study/}{0}{40}
    \caption{Visual comparison of joint object and camera motion control.}
    \label{fig:eval-shot-design}
\end{figure}

%% file: Tables/tab-ablation-study.tex

\begin{table}[t]
  \caption{Ablation study of applying different camera motion representations on our base model. $^*$: Zero-shot performance.
  }
  \label{tab:ablation_study_camera}
\resizebox{\linewidth}{!}{
  \setlength{\tabcolsep}{2.1pt}
  \centering
  \begin{tabular}{lcccccc}
  \toprule
    Variant  &RotErr$\downarrow$ &TransErr$\downarrow$ &FVD$\downarrow$ &+ Tokens(\%)$\downarrow$& Latency$\downarrow$\\
    \midrule
Gaussian map$^*$  & 0.8250& 0.2551& 116.47&99.7 &75s\\
 Plucker  & \best{0.5965}& 0.2244& \best{25.71}& 319.2&210s\\
 \rowcolor{Gray}
 Traj. coeff. (Ours)$^*$ & 0.6334& \best{0.2188}&34.09& \best{1.1}& \best{32s}\\
  \bottomrule
  \end{tabular}
}
\end{table}

%% file: Figures/fig-eval-camera-ablation.tex
\begin{figure*}[htbp]
    \centering
    \animategraphics[autoplay,loop,width=0.98\linewidth, trim=0 0 35 0, clip]{24}{Frames/app-camera-motion-ablation/}{0}{30} 
    \caption{Visual comparison of the results generated by different variants of our method.}
    \label{fig:eval-camera-ablation}
\end{figure*}

%% file: Sections/sec-6-conclusion.tex
\section{Conclusion}
\label{sec-6-conclusion}


We presented \MOCA, a unified I2V synthesis system that enables cinematic shot design with flexible control of camera and object motion. Using a Motion Signal Translation module, 
\MOCA converts intuitive 3D motion planning into precise 2D control signals for training video models without relying on 3D annotations, and hence broadens the source of training data. Comprehensive evaluations showed \MOCA's effectiveness in generating diverse, high-quality videos that faithfully reflect user motion intent.

%% file: Sections/X_supp.tex
\clearpage
\newcommand{\nocontentsline}[3]{}
\newcommand{\tocless}[2]{\bgroup\let\addcontentsline=\nocontentsline#1{#2}\egroup}

\newcommand{\Appendix}[1]{
  \refstepcounter{section}
  \section*{Appendix \thesection:\hspace*{1.5ex} #1}
  \addcontentsline{toc}{section}{Appendix \thesection}
}
\newcommand{\SubAppendix}[1]{\tocless\subsection{#1}}
\maketitlesupplementary
\appendix

\tableofcontents
\addtocontents{toc}{}

Please check our project page \url{https://motion-canvas25.github.io/} for video results.

\section{Additional Details}
\label{sec:add_details}
Our MotionCanvas model is trained using 16 nodes of NVIDIA H100 (80GB) GPUs (8 GPUs on each node). On a single H100 GPU, generating a 32-frame video clip at a resolution of 352×640 with 50 denoising steps takes approximately 32 seconds. During training, we exclusively optimize the DiT transformer blocks and the additional Linear/MLP layers introduced for bounding box conditioning and DCT coefficient tokenization, while keeping all other modules frozen.

\begin{figure*}[htbp]
    \centering
    \includegraphics[width=0.85\linewidth]{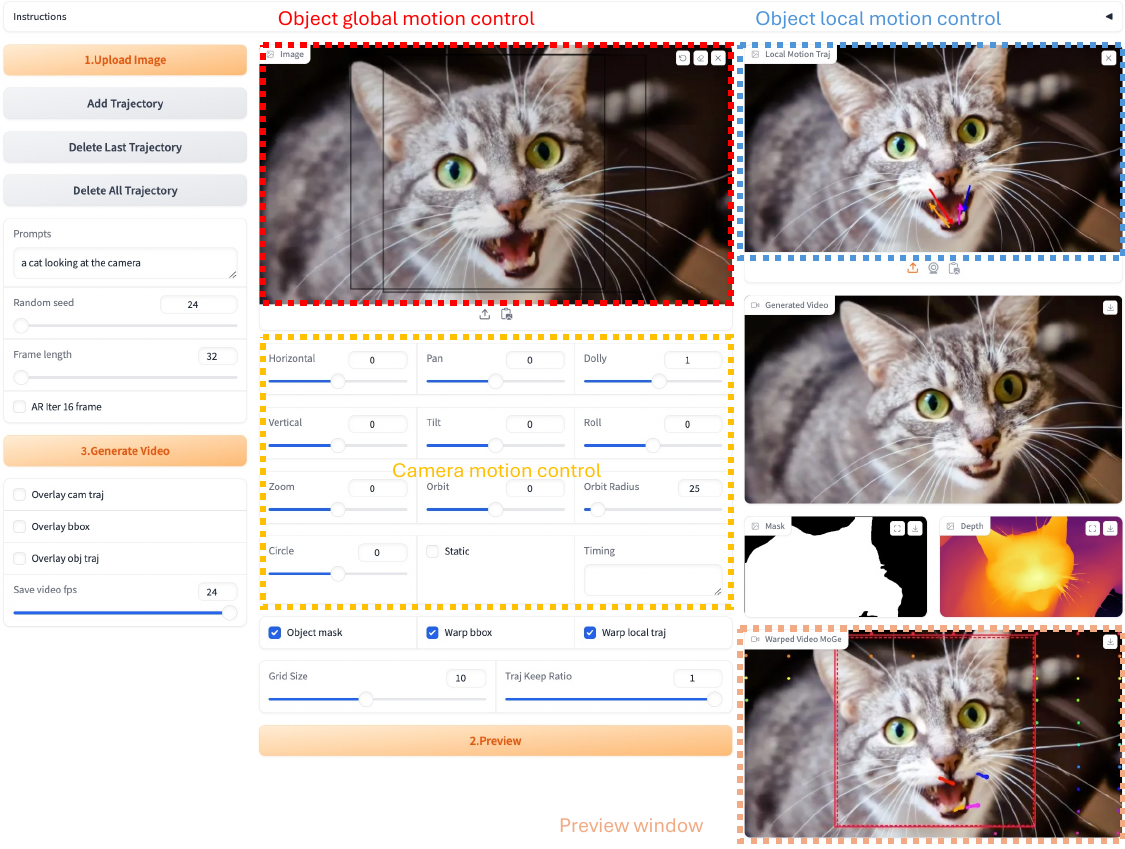} 
    \caption{A sample of the designed user interface for our MotionCanvas.}
    \label{fig:ui}
\end{figure*}
\section{User Interface}
\label{sec:ui}
We designed a sample user interface to provide flexible and interactive control over camera motion, object global and local motion, and their timing. An example of the user interface is illustrated in Fig.~\ref{fig:ui}.

\textbf{Specifying Camera Motion.} To facilitate a user-friendly approach for defining camera motion trajectories, the interface allows users to combine $M$ base motion patterns with configurable parameters such as direction (positive or negative) and speed (absolute value), as demonstrated in the ``Camera Motion Control" panel in Fig.~\ref{fig:ui}. Specifically, the base motion patterns include:

\begin{itemize}[left=2em]
    \item Horizontal (Trucking) left/right
    \item Panning left/right
    \item Dolly in/out
    \item Vertical (Pedestal) up/down
    \item Tilt up/down
    \item Roll clockwise/anti-clockwise
    \item Zoom in/out
    \item Orbit left/right (adjustable radius)
    \item Circle clockwise/anti-clockwise
    \item Static
\end{itemize}

The sign (positive or negative) and the absolute value of the number associated with each motion pattern define the corresponding camera poses relative to the zero pose at the first frame (i.e., translation and rotation vectors).

\textbf{Specifying Scene-aware Object Global Motion.} To enable user control over object global motion trajectories, we provide an interactive canvas (see Fig.~\ref{fig:ui}, ``Object Global Motion Control") where users can draw starting and ending bounding boxes, as well as optional intermediate points. A smooth bounding box trajectory can be obtained by applying Catmull-Rom spline interpolation. For each bounding box, users can optionally specify a reference depth point on the image. Additionally, users can decide whether to directly use this scene-space bounding box sequence as a condition. Utilizing the scene-space bounding box is particularly more effective for creating cinematic effects such as ``follow shots" or ``dolly shots".

For standard scene-aware object global motion control, the scene-space bounding boxes are assigned depth values, as described in Section~3.2 of the main paper. The bounding box sequence is then converted into screen space using the proposed Motion Signal Translation module.

\textbf{Specifying scene-aware object local motion.} We also provide a dedicated canvas for controlling object local motion (see Fig.~\ref{fig:ui}, ``Object Local Motion Control"). Users can draw any number of point trajectories, which are assigned depth values as outlined in Section 3.2 of the main paper. Similar to bounding boxes, these point trajectories are transformed into screen space based on the camera motion and object global motion. The object global motion transformation takes effect only when the starting point of the trajectory lies within the object's semantic region. We then transform the object's local motion point trajectories by maintaining their relative positions with respect to the underlying bounding box.

Additionally, our user interface includes a ``Preview Window" that allows users to visualize the generated videos, as well as the bounding box sequences and point trajectories in both scene space and screen space.

\begin{figure}[htbp]
    \centering
    \includegraphics[width=1\linewidth]{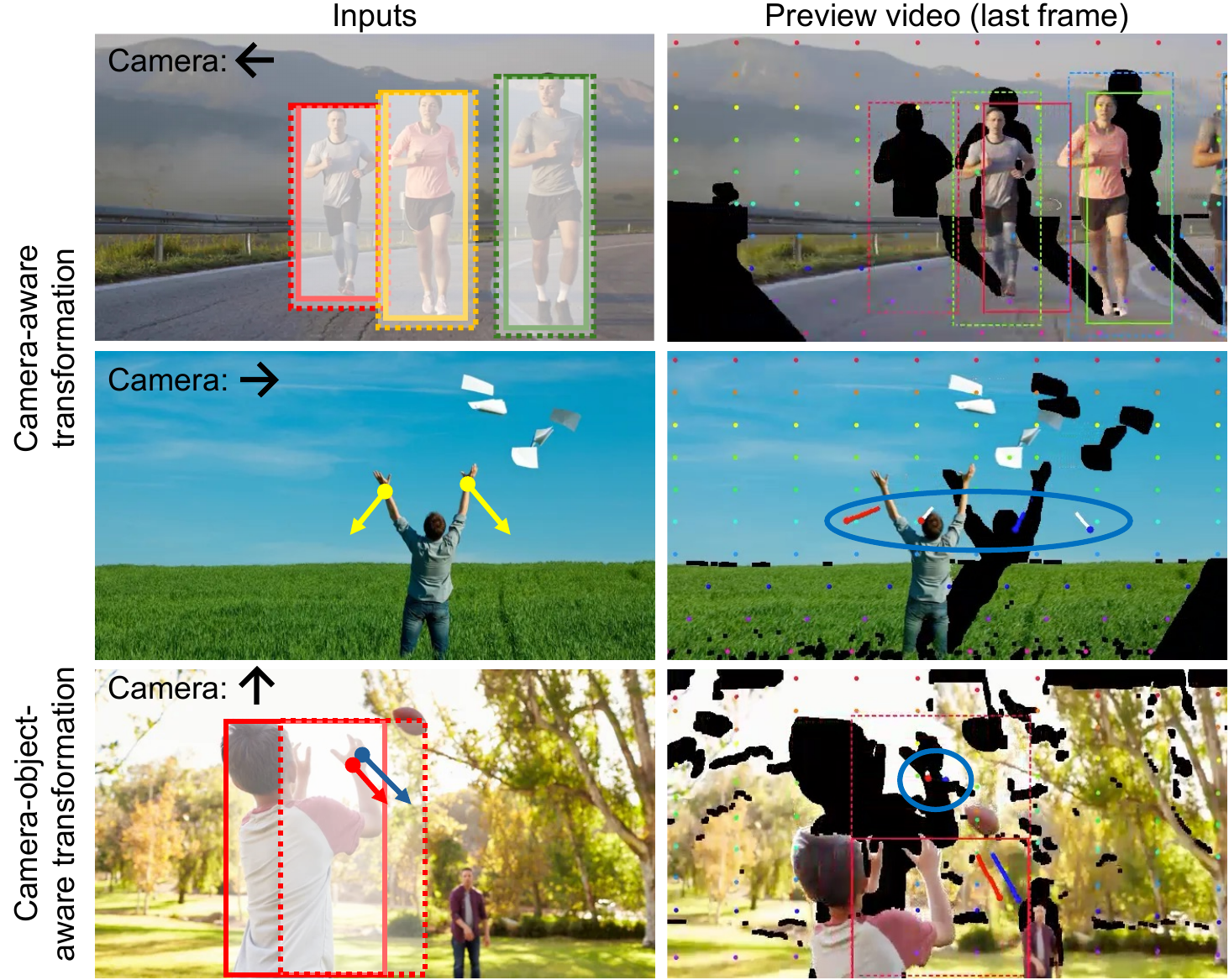} 
    \caption{Illustration of camera-aware transformation and camera-object-aware transformation. In preview videos, dash-line bounding boxes represent the scene-space inputs, while the solid ones with the same color denote the transformed screen-space motion signals. Similarly, point trajectories with white trace indicate scene-space user motion design, while colored ones represent transformed signals. Better investigation in supplementary videos.}
    \label{fig:transform}
\end{figure}
\section{Essentiality of Camera-aware and Camera-object-aware Transformations}
\label{sec:transform}
Drawing inspiration from classical graphics imaging techniques, we introduce a Motion Signal Translation module to convert scene-space user-defined motion intents into screen-space motion signals. This enables joint control of camera and object motions in a 3D-aware manner for image-to-video generation. The Motion Signal Translation module incorporates a hierarchical transformation framework that accounts for the intertwining nature of camera and object motions. To illustrate the effectiveness of these transformations, we provide visual comparisons highlighting both camera-aware and camera-object-aware transformations.

\textbf{Camera-aware Transformation for Object Global or Local Motion.}
First, we present the camera-aware transformation for object global motion control in Fig.~\ref{fig:transform}(top). In the preview video (last frame), the dashed-line bounding boxes represent the scene-space inputs specified by the user, while the solid bounding boxes of the same color denote the corresponding transformed screen-space motion signals.

In this example, people are running forward on the road as controlled by the user’s input (bounding boxes). When a trucking-left camera motion is applied, all the people should naturally move to the right on the screen. Using our camera-aware transformation, the screen-space object bounding boxes are correctly calibrated, ensuring that the resulting animation appears accurate and more natural (refer to the supplementary webpage: ``Additional Analysis -- Essentiality of Camera-aware and Camera-object-aware Transformations").

A similar conclusion holds for the camera-aware transformation applied to object local motion, as shown in Fig.~\ref{fig:transform}(middle).

\textbf{Camera-object-aware Transformation for Object Local Motion.}
Camera-object-aware transformation implies that translating scene-space point trajectory specifying the object local motion to screen-space signals must take into account both the camera motion and object global motion.  For example, as illustrated in Fig.~\ref{fig:transform}(bottom), the trajectory of an object’s local motion, such as ``putting down hands" must account for both the body’s movement and the camera’s pedestal-up motion. As demonstrated in ``Additional Analysis -- Essentiality of Camera-aware and Camera-object-aware Transformations", our transformation produces a life-like and accurate video, whereas the variant without these transformations results in unnatural and incorrect motion.

\begin{figure}[t]
    \centering
    \includegraphics[width=1\linewidth]{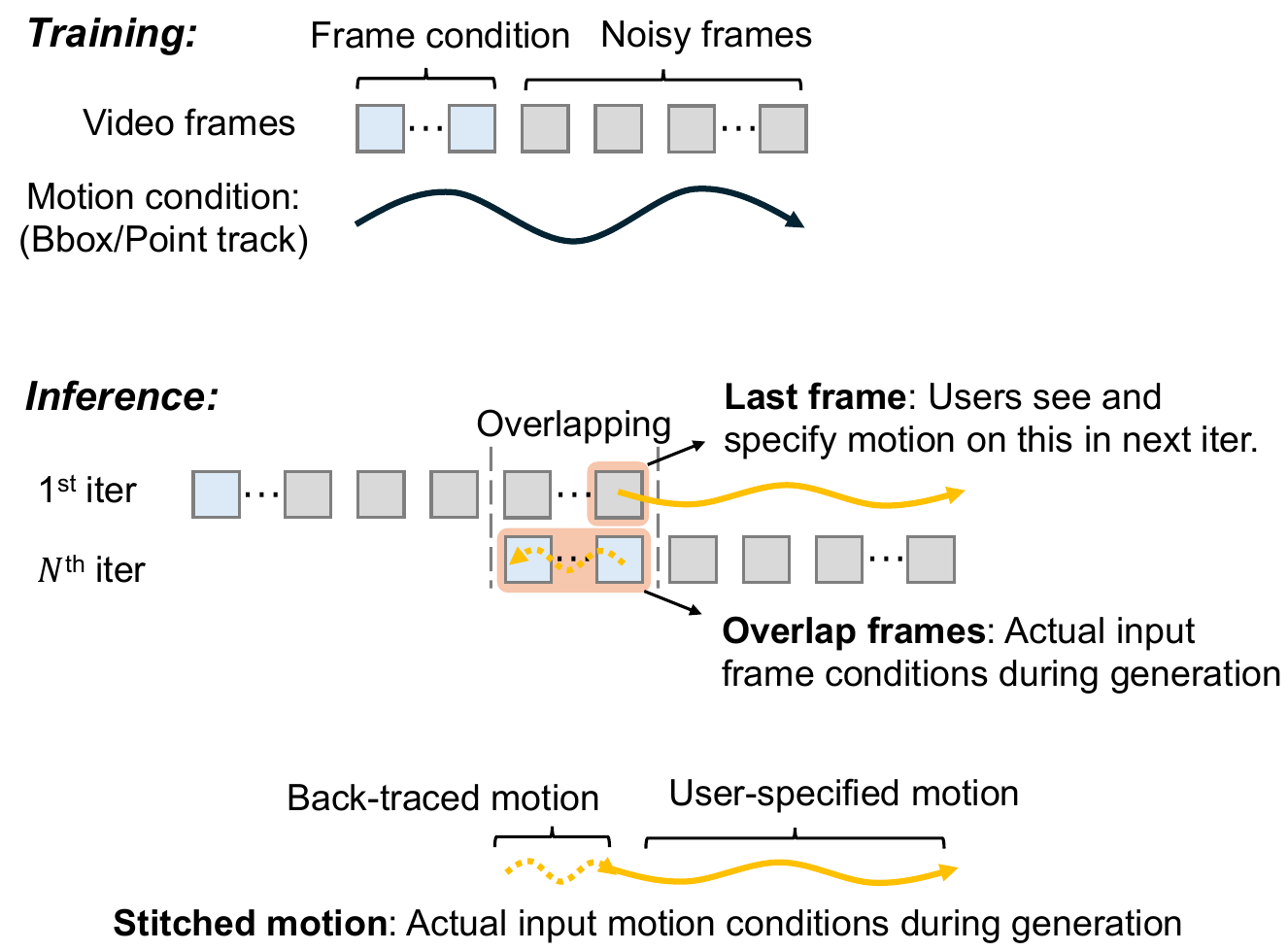} 
    \caption{Illustration of the recomputation process for input motion conditions in our MotionCanvas$_\text{AR}$ during inference.}
    \label{fig:back_trace}
\end{figure}

\section{More Details of MotionCanvas$_\text{AR}$}
\label{sec:ar}
To enhance support for long video generation and address motion discontinuities, we introduce a 16-frame conditioned 64-frame MotionCanvas$_\text{AR}$, designed to generate videos in an auto-regressive manner. This model builds on our 32-frame motion-conditioned I2V model (single-frame-conditioned) and is fine-tuned for an additional 120K iterations, while retaining the same training configurations.

To further refine the input motion signals and better align them with the training setup, we recompute the screen-space motion signals by integrating the user’s motion intent with back-traced motions, as illustrated in Fig.~\ref{fig:back_trace}. This method ensures smoother, more consistent motion generation throughout the video.

\section{User Study}
\label{sec:user}
The designed user study interface is shown in Figure~\ref{fig:user_study_screen_shot}. We collect 15 representative image cases from the Internet and design various motion controls. We then generate the video clips results by executing the official code~\cite{wu2025draganything,niu2025mofa}. For the user study, we use these video results produced by shuffled methods based on the same set of input conditions. In addition, we standardize all the produced results by encoding FPS$=$8 for 14 generated frames, yielding $\sim$2-second videos for each method. This process ensures a fair comparison.

The user study is expected to be completed with 7--15 minutes (15 cases $\times$ 3 sub-questions $\times$ 10--20 seconds for each judgement). To remove the impact of random selection, we filter out those comparison results completed within three minutes. For each participant, the user study interface shows 15 groups of video comparisons, and the participant is instructed to evaluate the videos for three times, \ie, answering the following questions respectively: (i) ``Which one shows the best motion adherence?"; (ii) ``Which one has the best motion/dynamic quality?"; (iii) ``which one shows the best frame fidelity?". Finally, we received 35 valid responses from the participants.

\begin{figure}[t]
    \centering
    \includegraphics[width=1.0\linewidth]{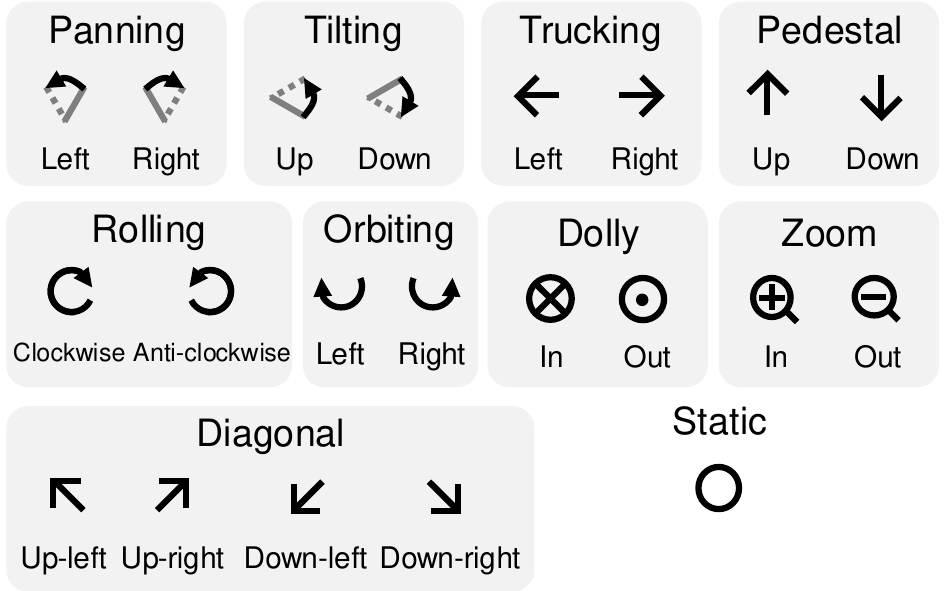} 
    \caption{Legend of camera motions used in the main paper.}
    \label{fig:legend}
\end{figure}

\begin{figure*}[htbp]
    \centering
    \includegraphics[width=0.8\linewidth]{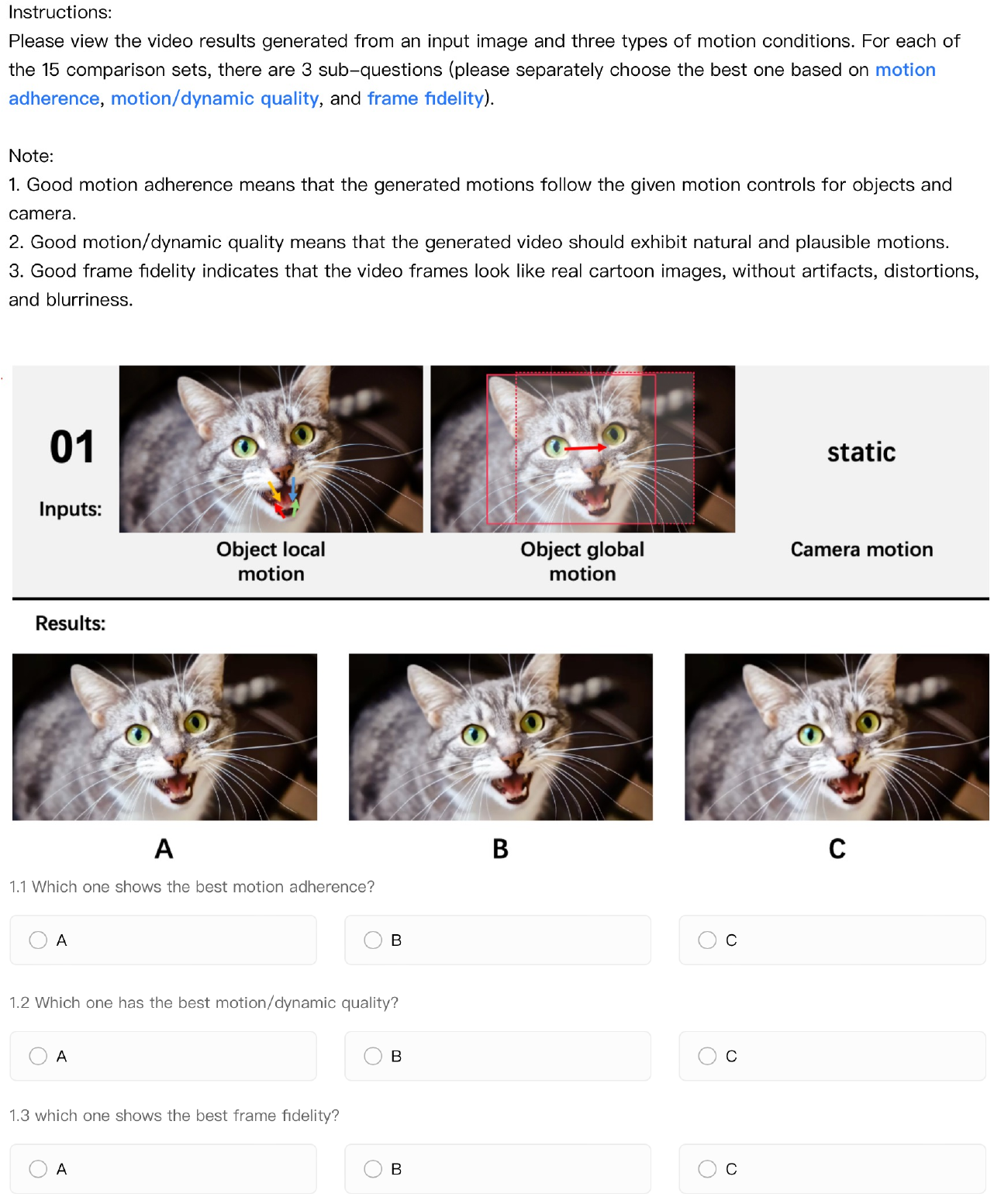} 
    \caption{Designed user study interface. Each participant is required to evaluate 15 video comparisons and respond to three corresponding sub-questions for each comparison. Only one video is shown here due to the page limit.}
    \label{fig:user_study_screen_shot}
\end{figure*}

\section{Additional Analysis}
\label{sec:add_analysis}
\subsection{Effect of Point Track Density on Camera Motion Control}

We investigate the effect of point track density on camera motion control by specifying orbit right camera motion with different numbers of 2D point tracks. The visual comparison result is shown in the supplementary webpage `Additional Analysis -- Effect of Point Track Density on Camera Motion Control'. As can be seen, the motion is underspecified with significant ambiguity when providing low-density tracks. Hence, the generated camera motion does not follow the control and tends to be trucking left. By providing higher-density tracks, the generated motion can better adhere to the orbit camera motion.

\subsection{Effect of Text Prompt}
We use simple text descriptions throughout all experiments. To further investigate the effect of text prompts on our MotionCanvas, we show the visual comparison of gradually more detailed text prompts in the supplementary webpage `Additional Analysis-- Effect of Text Prompt.'. It demonstrate that text prompt does not have a significant effect on the camera motion control. However, it can generate diverse dynamics like `raining' and `turning around'.

\section{Limitations and Future Work}
\label{sec:limitations}

Our work introduces a novel framework for enhancing I2V with holistic motion controls, enabling cinematic shot design from a single image. While this paper made substantial progress on toward this important application, challenges remain, opening up opportunities for future research. 

First, our use of a video diffusion model (VDM) for the video synthesis module, while enabling high-quality video generation, results in relatively slow inference times (approximately 35 seconds for a 2-second video). This computational cost, typical of modern VDMs, currently limits real-time applications. Exploring alternative, more efficient generative models is a promising direction for future work. 

Second, our current method approximates object local motion by assuming each object lies on a frontal parallel depth plane. Although effective for most natural scenes as the depth variation within the object are typically small compared to object's distance to the camera, this pseudo-3D approximation may not be suitable for extreme close-up or macro shots where depth variations within the object are significant. In future work, it will be interesting to investigate integrating more explicit 3D formulation for handling such scenarios. 

Finally, our system currently does not explicitly constrain the harmonization between motion design and textual prompts. On one hand, this offers the flexibility for users to leverage both control modalities jointly to explore creative variations. On the other hand, this leaves the possibility of conflicting motion signals between the modalities. For example, in Fig. 8 (top row) in the main paper, while the text prompt indicates the cat to be waiting instead of moving, when motion design explicitly control the cat to moves forward in the later part of the videos (note that such global object control was used when generated those results but was not illustrated in the figure to avoid cluttered visualization), such motion control overrides the ones hinted in the textual prompts. Explicit motion-aware prompt harmonization can be a fruitful research direction to extend our work.  

\section{Camera Motion Legend}
\label{sec:legend}
The legend of camera motions used in the main paper is presented in Fig.~\ref{fig:legend}.